\documentclass[letterpaper, 10 pt, conference]{ieeeconf}  

\IEEEoverridecommandlockouts                              

\usepackage{amsmath} 
\usepackage{amssymb}  

\usepackage{hyperref}
\usepackage{url}
\usepackage{cite}
\usepackage{subfigure}
\usepackage{graphicx}
\usepackage{amsfonts}

\usepackage{algorithm}
\usepackage{algpseudocode}
\usepackage[dvipsnames]{xcolor}

\newcommand{\cmt}[1]{}

\long\def\ignorethis#1{}

\newcommand{\etal}{{\em{et~al.}\ }}
\newcommand{\eg}{e.g.\ }
\newcommand{\ie}{i.e.\ }

\newcommand{\vc}[1]{\ensuremath{\mathbf{#1}}}

\newcommand{\pctab}{\hspace{0.2in}}

\title{\LARGE \bf
Sim-to-Real Transfer for Biped Locomotion
}

\author{Wenhao Yu$^{1}$, Visak CV Kumar$^{1}$, Greg Turk$^{1}$, C. Karen Liu$^{1}$
\thanks{$^{1}$Authors are with School of Interactive Computing, Georgia Institute of Technology, Atlanta, GA, 30308
        {\tt\small wenhaoyu@gatech.edu, visak3@gatech.edu, karenliu@cc.gatech.edu, turk@cc.gatech.edu}}%
}

\begin{document}

\maketitle
\thispagestyle{empty}
\pagestyle{empty}

\begin{abstract}

We present a new approach for transfer of dynamic robot control policies such as biped locomotion from simulation to real hardware.  Key to our approach is to perform system identification of the model parameters $\boldsymbol{\mu}$ of the hardware (e.g. friction, center-of-mass) in two distinct stages, before policy learning (pre-sysID) and after policy learning (post-sysID).  Pre-sysID begins by collecting trajectories from the physical hardware based on a set of  generic motion sequences. Because the trajectories may not be related to the task of interest, pre-sysID does not attempt to accurately identify the true value of  $\boldsymbol{\mu}$, but only to approximate the range of $\boldsymbol{\mu}$ to guide the policy learning. Next, a Projected Universal Policy (PUP) is created by simultaneously training a network that projects $\boldsymbol{\mu}$ to a low-dimensional latent variable $\boldsymbol{\eta}$ and a family of policies that are conditioned on $\boldsymbol{\eta}$. The second round of system identification (post-sysID) is then carried out by deploying the PUP on the robot hardware using task-relevant trajectories. We use Bayesian Optimization to determine the values for $\boldsymbol{\eta}$ that optimize the performance of PUP on the real hardware.  We have used this approach to create three successful biped locomotion controllers (walk forward, walk backwards, walk sideways) on the Darwin OP2 robot.

\end{abstract}

\section{Introduction}

Designing locomotion controllers for biped robots is a challenging task that often depends on tuning the control parameters manually in a trial-and-error fashion. Recent advancements in Deep Reinforcement Learning (DRL) have shown a promising path in automating the design of robotic locomotion controllers \cite{peng2018deepmimic,schulman2017proximal,YuSIGGRAPH2018,schulman2015trust}. However, the amount of training samples demanded by most DRL methods is infeasible to acquire from the real world for high-risk tasks, such as biped locomotion. While computer simulation provides a safe and efficient way to learn motor skills, a policy trained in simulation does not often transfer to the real hardware due to various modeling discrepancy between the simulated and the real environments, referred to as the Reality Gap \cite{neunert2017off}.

To overcome the reality gap, recent work has investigated more sophisticated system identification procedures that improve the accuracy of the model, developed more robust control policies that work for a large variety of simulated environments (\ie domain randomization), or interleaved system identification with robust policy learning in an iterative algorithm. While these approaches have shown impressive results on manipulation and quadruped locomotion tasks, it is not clear whether this success can be extended to biped locomotion. The inherent instability of biped locomotion makes standard system identification ineffective because it is challenging to collect task-relevant data for system identification prior to policy training. Without an effective system identification procedure, robust policy learning would only work if the default models are reasonably accurate for approximating the robot dynamics, sensors, and actuators. Furthermore, a balanced biped locomotion policy is generally more susceptive to discrepancies between the training and testing environments, rendering the domain randomization technique alone insufficient to conquer the reality gap.

\begin{figure}[t!]
\centering
\subfigure{\includegraphics[width=0.8\linewidth]{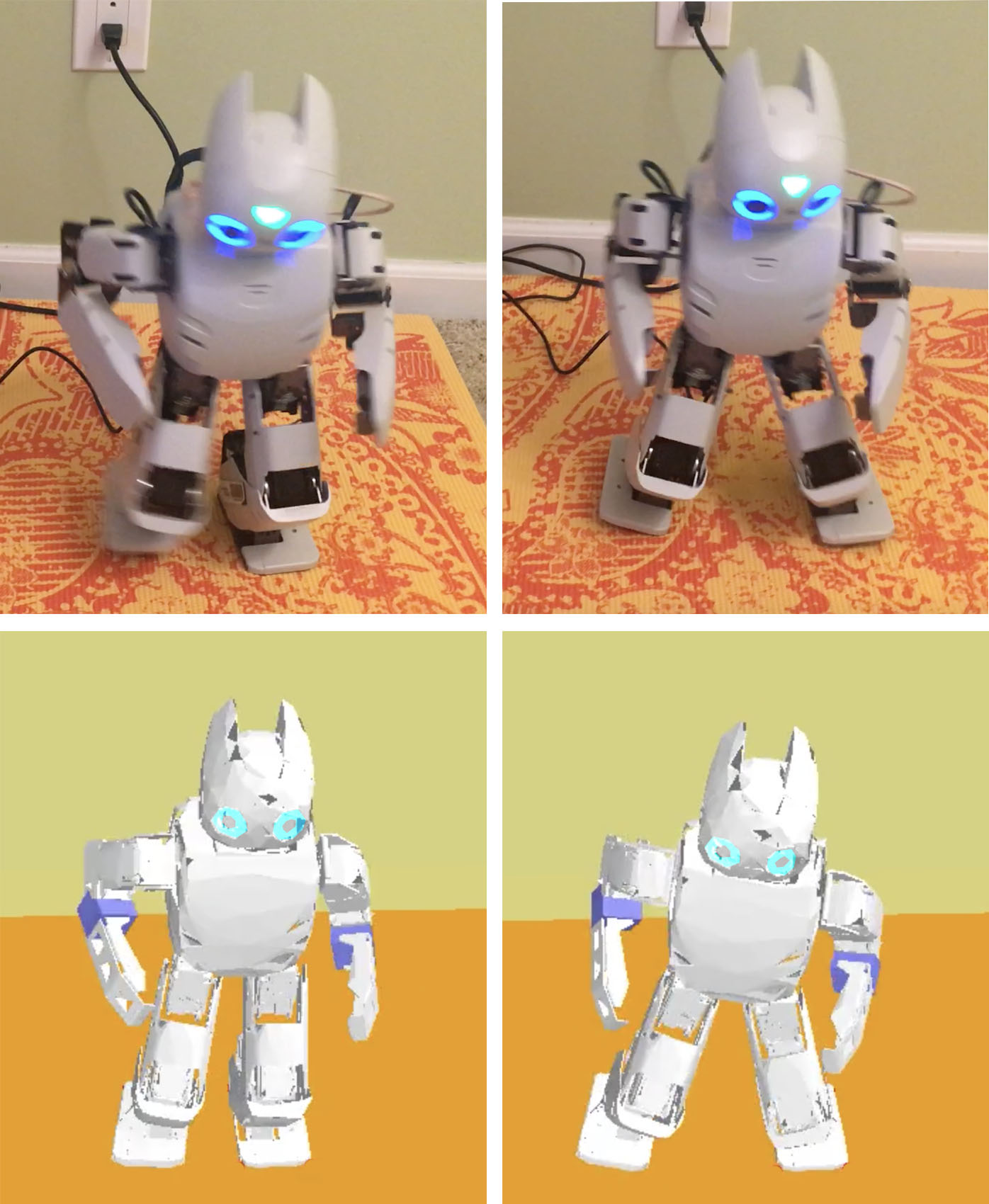}}
\caption{The simulated and the real Darwin OP2 robot trained to walk sideways.}
\vspace{-3mm}
\label{fig:side_sim2real}
\end{figure}

\begin{figure}[t!]
\centering
\subfigure{\includegraphics[width=\linewidth]{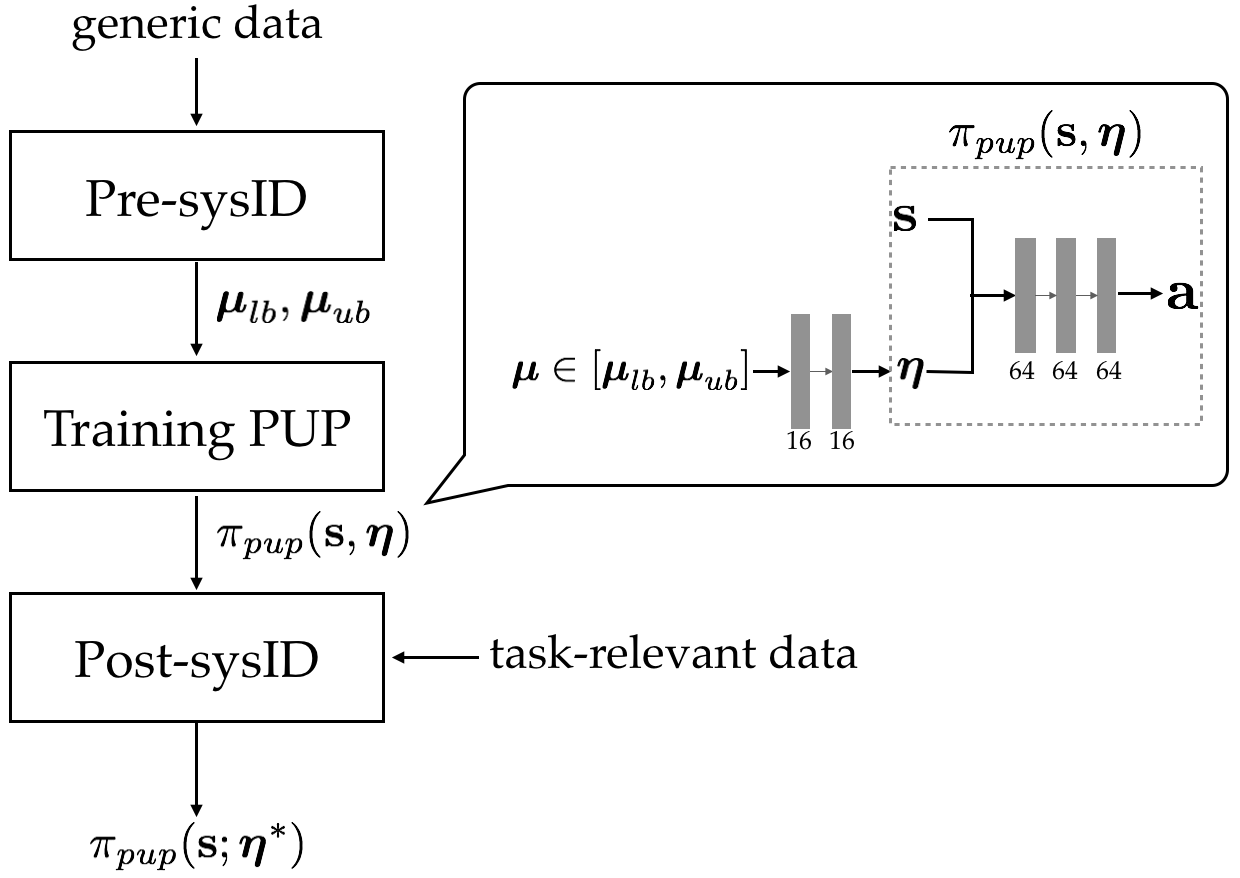}}
\vspace{-2mm}
\caption{Overview of the proposed algorithm.}
\label{fig:overview}
\vspace{-3mm}
\end{figure}

This paper aims to develop a biped locomotion controller by training a policy in simulation and deploying it to consumer-grade robotic hardware (\eg Darwin OP2 which costs less than \$10,000 USD). The key idea of our approach is to split the system identification procedure into two stages: one prior to the policy training (pre-sysID) and one subsequent to the policy training (post-sysID) (Figure \ref{fig:overview}). Since the real-world data collected prior to the policy training are usually irrelevant to the task, the goal of pre-sysID is not to accurately identify the true value of model parameters, but only to approximate the range of model parameters in order to train a policy later. After the policy is trained in simulation, the post-sysID is then able to use task-relevant data to identify the model parameters that optimize the performance of the policy deployed in the real world. The critical component that bridges the two system ID processes is the Projected Universal Policy (PUP). We simultaneously train a network that projects the model parameters to a low-dimensional latent variable, together with a family of policies that are conditioned on the latent space. As such, PUP is a policy which can be modulated by a low-dimensional latent variable to adapt to any environment within the range of model parameters it is trained for. Once PUP is well trained in simulation, we only need to perform one more system identification, \ie post-sysID, to optimize the transfer performance of PUP, without the need of iterating between system identification and policy learning.

We demonstrate our algorithm on training locomotion controllers for the Darwin OP2 robot to perform forward, backward and sideway walks. Our algorithm can successfully transfer the policy trained in simulation to the hardware in $25$ real-world trials. We also evaluate the algorithm by comparing our method to two baseline methods: 1) identify a single model during system identification and train a policy for that model, and 2) use the range of parameters from pre-sysID to train a robust policy.

\section{Related Work}


Transferring control policies from a source domain to a different target domain has been explored by a number of research groups. In the context of reinforcement learning, a comprehensive survey on transfer learning was reported by Taylor and Stone \cite{Taylor2009}. One notable application of transfer learning in reinforcement learning is sim-to-real transfer, where policies trained in simulation are transferred to real robots. Sim-to-real transfer allows automatic training of robot controllers in a safer environment, however, it is a challenging problem due to the presence of the Reality Gap \cite{neunert2017off}. A few recent works have shown successful transfer of policies between simulated environments \cite{yu2018policy,YuRSS17,rajeswaran2016epopt,HA2015,Mandlekar,Gupta2017, pinto2017robust}, however, these were not tested on real robotic systems.


Among methods that successfully transfer control policies to real robots, two key ideas are instrumental: system identification \cite{Ljung1998} and robust policy generation. System identification aims to bridge the reality gap by identifying simulation models that can accurately capture the real-world events. Researchers have shown that, with carefully identified simulation models, it is possible to transfer learned trajectories or policies to real hardware \cite{TanRSS18,tan2016simulation,Abbeel2005, deisenroth2011pilco, hwangbo2019learning, Park, Golemo2018}. For example, Tan et al. \cite{TanRSS18} combined domain randomization with accurate identification of the servo motor model and demonstrated successful transfer of locomotion policies for a quadruped robot. The key idea in their work is to identify a non-linearity current-torque relationship in motor dynamics. Hwangbo et al. \cite{hwangbo2019learning}, trained a deep neural network that maps from motor commands to torques using data from the real actuators of a quadruped robot. The trained model then replaces the motor model in the simulator, which is used to train control policies for the quadruped. They demonstrated transferring of agile and dynamic motions on a real robot. In our work, we also utilize a neural network for modeling the motor dynamics, however, our method do not rely on the high-end actuators to generate ground-truth torque data as in \cite{hwangbo2019learning}.

The other key component in sim-to-real transfer is training robust policies. To make a policy more robust, numerous approaches have been explored such as adding adversarial perturbations during training in Pinto et al. \cite{pinto2017robust}, using ensemble of models \cite{Mordatch,Lowrey} randomizing sensor noise in Jakobi et al. \cite{Noise} and domain randomization \cite{tobin2017domain,peng2018sim,learndexmanipulation18}. In Peng et al. \cite{peng2018sim}, an LSTM policy was trained with dynamics randomization and transferred to a real robotic manipulator. Andrychowicz et al. \cite{learndexmanipulation18} demonstrated dexterous in-hand manipulation on an anthropomorphic robot hand, the Shadow Dexterous Hand, by introducing randomization in both perception and dynamics of the simulation during training. Although these methods have shown promising results, they in general assume that the dynamics of the testing environment is not too far away from the set of training environments. One may increase the range of training environments to avoid this, however, it will require more samples and compute for training the policy and may lead to overly-conservative behaviors.



Another approach in sim-to-real transfer is to further fine-tune a trained policy using data collected from the real hardware \cite{rusu2016sim,Chebotar,chen2018hardware}. Chebotar et al.\cite{Chebotar} presented a technique to interleave system identification and policy learning for manipulation tasks. At each iteration, the distribution of the randomized parameters was optimized by minimizing the difference between the trajectories collected in simulation and real-world. They demonstrated transfer for tasks such as draw opening. However, it is unlikely to work for bipedal locomotion tasks because the quick failure of the policy on the robot during initial iterations may not provide enough information for meaningful updates to the parameter distribution. The closest work to ours is that of Cully et al. \cite{cully2015robots}. Their algorithm searches over the space of previously learned behaviours that can compensate for changes in dynamics, like a damage to the robot. The key difference is that we generate a parameterized family of policies by varying system dynamics during policy training, while they manually specify variations in the gaits and use trajectory optimization to obtain a discrete behaviour space.


Despite concerns about safety and sample complexity of DRL methods, there has been success in directly training locomotion controllers on the real robot \cite{Haarnoja, Ha}. In Ha et al. \cite{Ha}, a policy was directly trained on a multi-legged robot. The training was automated by a novel resetting device which was able to re-initialize the robot during training after each rollout. In Haarnoja et al. \cite{Haarnoja}, a policy was trained for a real quadruped robot in under two hours from scratch using soft-actor critic algorithm \cite{SAC}. Despite these success in learning legged locomotion tasks, directly training policies on a biped robot is still challenging due to the frequent manual resetting required during training and the potential safety concern from the inherent instability.



\section{Pre-training System Identification (Pre-sysID)}

\label{sec:presysID}

The goal of a standard system identification procedure is to tune the model parameters $\boldsymbol{\mu}$ (e.g. friction, center of mass) such that the trajectories predicted by the model closely match those acquired from the real-world.  One important decision in this procedure is the choice of data to collect from the real world. Ideally, we would like to collect the trajectories relevant to the task of interest. For biped locomotion, however, it is challenging to script successful locomotion trajectories prior to the policy training. Without task-relevant trajectories, any other choice of data can become a source of bias that may impact the resulting model parameters $\boldsymbol{\mu}$. Our solution to this problem is that, instead of solving for the optimal set of model parameters, Pre-sysID only attempts to approximate a reasonable range of model parameters for the purpose of domain randomization during policy learning. As such, we can use less task-relevant trajectories to cover a wide range of robot behaviors that may be remotely related to the task. In the case of locomotion, we use two set of trajectories for system identification: joint exercise without contact and standing/falling with ground contact. We use a set of pre-scripted actions to create these trajectories. (See details in Section \ref{sec:exp_setup}). 

\subsection{Optimizing Range of Model Parameters}
We optimized the model parameters $\boldsymbol{\mu}$ by creating simulated trajectories using the same pre-scripted actions that were used to collect the real-world trajectories.  The fitness of a given $\boldsymbol{\mu}$ is given by the deviation between these simulated and real-world trajectories. Instead of trying to find a single simulation model that perfectly explains all the training data, we optimize for a set of models simultaneously, each of which fits a subset of the training trajectories. Specifically, we first use the entire set of trajectories to optimize a nominal set of model parameters $\hat{\boldsymbol{\mu}}$. We then select random subsets of the training trajectories, for each subset we optimize the model parameters again with $\hat{\boldsymbol{\mu}}$ as initial guess. During the optimization of each subset, we add a regularization term $w_{reg}||\boldsymbol{\mu}-\hat{\boldsymbol{\mu}}||^2$ to the objective function so that $\boldsymbol{\mu}$ will not go to local minima that are far away. We use $w_{reg}=0.05$ in our experiments. This results in a set of optimized simulators, each of which can better reproduce a subset of the training data than $\hat{\boldsymbol{\mu}}$. We then extract the range of the simulation parameters by taking the element-wise maximum and minimum of the optimized $\boldsymbol{\mu}$'s and expand them by $10\%$ to obtain the bounds $\boldsymbol{\mu}_{lb}$ and $\boldsymbol{\mu}_{ub}$.

We use CMA-ES \cite{hansen1995adaptation}, a sampling-based optimization algorithm, to optimize $\boldsymbol{\mu}$. To evaluate the fitness of a sampled $\boldsymbol{\mu}$, we compare the trajectories generated by the simulation to those from the hardware:
\begin{align}
    \label{eq:sysid_loss}
    L = & \frac{1}{|D|}\sum_{D} \sum_t \|\bar{\mathbf{q}}_t - \mathbf{q}_t\| \\ & + \frac{10}{|D_{s,f}|}\sum_{D_{s,f}} \sum_t \|\bar{\mathbf{g}}_t - \mathbf{g}_t\|\nonumber \\ & +
    \frac{20}{|D_{s}|}\sum_{D_{s}} \sum_t \| \Delta C^{feet}_t \|, \nonumber
\end{align}
where $D$ denotes the entire set of input training trajectories from hardware, $D_{s,f}$ denotes the subset of standing or falling trajectories, and $D_s$ contains only the standing trajectories. The first term measures the difference in the simulated motor position $\mathbf{q}$ and the real one $\bar{\mathbf{q}}$. The second term measures the difference in the roll and pitch of the robot torso between the simulated one $\mathbf{g}$ and the real one $\bar{\mathbf{g}}$. The third term measures the movement of the feet in simulation since the foot movement in the real trajectories is zero for those in $D_s$.

\begin{algorithm}[t]
\caption{System Identification of Parameter Bounds}\label{alg:sysid}
\begin{algorithmic}[1]
\State Collect trajectories on hardware and store in $D$\;
\State $\hat{\boldsymbol{\mu}} = \arg\min_{\boldsymbol{\mu}} L(D, \boldsymbol{\mu})$ \;
\For{$i=1:N$}
\State $D_i \leftarrow$ random subset of $D$ \;
\State $\boldsymbol{\mu}_i$ = $\arg\min_{\boldsymbol{\mu}} L(D_i, \boldsymbol{\boldsymbol{\mu}}) + w_{reg}\|\boldsymbol{\mu} - \boldsymbol{\hat{\mu}}\|^2$ \;
\EndFor
\State $\boldsymbol{\mu}_{max}$, $\boldsymbol{\mu}_{min}\leftarrow$ per-dimension max and min of $\boldsymbol{\mu}_i$\;
\State $\boldsymbol{\mu}_{lb} = \boldsymbol{\mu}_{min} - 0.1 (\boldsymbol{\mu}_{max}-\boldsymbol{\mu}_{min})$\;
\State $\boldsymbol{\mu}_{ub} = \boldsymbol{\mu}_{max} + 0.1 (\boldsymbol{\mu}_{max}-\boldsymbol{\mu}_{min})$\;
\State \textbf{return} $\boldsymbol{\mu}_{lb}$, $\boldsymbol{\mu}_{ub}$
\end{algorithmic}
\end{algorithm}

\subsection{Neural Network PD Actuator}
We model the biped robot as an articulated rigid body system with actuators at joints. For such a complex dynamic system, there are often too many model parameters to identify using limited amounts of real-world data. Among all the model parameters in the system, we found that the actuator is the main source of modeling error, comparing to other factors such as mass, dimensions, and joint parameters, similar to the findings in \cite{hwangbo2019learning}. Therefore, we augment the conventional PD-based actuator model with a neural network to increase the expressiveness of the model, which we name Neural Network PD Actuator (NN-PD). For each motor on the robot, the neural network model takes as input the difference between the target position $\bar{\theta}_t$ and the current position of the motor $\theta_t$, denoted as $\Delta \theta_t$, as well as the velocity of the motor $\dot{\theta}_t$, and outputs the proportional and derivative gains $k_p$ and $k_d$. Unlike the high-end actuators used in \cite{hwangbo2019learning}, the actuators on Darwin OP2 are not capable of accurately measuring the actual torque being applied. As a result, we cannot effectively train a large neural network that outputs the actual torque. In our examples, we use a neural network model of one hidden layer with five nodes using tanh activation, shared across all motors. This results in network weights of $27$ dimensions, which is denoted by $\boldsymbol{\phi} \in \mathbb{R}^{27}$.

We further modulate the differences among motors by grouping them based on their locations on the robot: $g \in \{$HEAD, ARM, HIP, KNEE, ANKLE$\}$. The final torque applied to the motor is calculated as:
\begin{equation}
    \tau = clip(\rho_g k_p(\boldsymbol{\phi}) \Delta \theta  - \sigma_g k_d(\boldsymbol{\phi}) \dot{\theta} , -\tilde{\tau}, \tilde{\tau}), \nonumber
\end{equation}
where the function $clip(x, b, u)$ returns the upper bound $u$ or the lower bound $b$ if $x$ exceeds $[b,u]$. Otherwise, it simply returns $x$. We define learnable scaling factors $\rho_g$ and $\sigma_g$ for each group, as well as a learnable torque limit $\tilde{\tau}$.

In addition to $\boldsymbol{\phi}$, $\sigma_g$ and $\tilde{\tau}$, our method also identifies the friction coefficient between the ground and the feet and the center of mass of the robot torso. Identifying the friction coefficient is necessary because the surface in the real world can be quite different from the default surface material in the simulator. We found that the CAD model of Darwin OP2 provided by the manufacturer has reasonably accurate inertial properties at each part, except for the torso where the on-board PC, sub-controller and $5$ motors reside. Thus, we include the local center of mass of the torso as an additional model parameter to identify.

The nominal model parameters $\hat{\boldsymbol{\mu}}$ we identify during pre-sysID include all the aforementioned parameters that has in total $41$ dimensions. However, we fix the motor neural network weights $\boldsymbol{\phi}$ and do not optimize the bounds for them. This is because neural network weights are trained to depend on each other and randomizing them independently might lead to undesired behavior. This results in the optimized parameter bounds $\boldsymbol{\mu}_{lb}, \boldsymbol{\mu}_{ub}$ to have dimension of $14$.

\section{Learning Projected Universal Policy}
We formulate the problem of learning locomotion controller as a Markov Decision Process (MDP), $(\mathcal{S}, \mathcal{A}, \mathcal{T}, r, p_0, \gamma)$, where $\mathcal{S}$ is the state space, $\mathcal{A}$ is the action space, $\mathcal{T}: \mathcal{S} \times \mathcal{A} \mapsto \mathcal{S}$ is the transition function, $r: \mathcal{S} \times \mathcal{A} \mapsto \mathbb{R}$ is the reward function, $p_0$ is the initial state distribution and $\gamma$ is a discount factor. The goal of reinforcement learning is to find a policy $\pi: \mathcal{S} \mapsto \mathcal{A}$, such that it maximizes the accumulated reward:
\begin{equation}
    J(\pi) = \mathbb{E}_{\mathbf{s}_0, \mathbf{a}_0, \dots, \mathbf{s}_T} \sum_{t=0}^{T} \gamma^t r(\mathbf{s}_t, \mathbf{a}_t),\nonumber
\end{equation}
 where $\mathbf{s}_0 \sim p_0$, $\mathbf{a}_t \sim \pi(\mathbf{s}_t)$ and $\mathbf{s}_{t+1}=\mathcal{T}(\mathbf{s}_t, \mathbf{a}_t)$.

Our method departs from the standard policy learning by training a Universal Policy (UP) $\pi_{up}:(\mathbf{s}, \boldsymbol{\mu}) \mapsto \mathbf{a}$ explicitly conditioned on the model parameters $\boldsymbol{\mu}$ \cite{YuRSS17}. Given the optimized bounds $\boldsymbol{\boldsymbol{\mu}}_{lb}$ and $\boldsymbol{\mu}_{ub}$ from Pre-sysID, UP can be viewed as a family of policies, each of which is trained for a particular environment, in which the transition function $\mathcal{T}_{\boldsymbol{\mu}}$ is parameterized by a particular $\boldsymbol{\mu}$ sampled from the uniform distribution $\mathcal{U}(\boldsymbol{\mu}_{lb}, \boldsymbol{\mu}_{ub})$. While UP has shown success for sim-to-sim transfer with similar dimensions of $\boldsymbol{\mu}$ in \cite{yu2018policy}, it poses great challenges for optimization in the real world on a biped robot (Section \ref{sec:post-sysid}).

To overcome the high dimensionality of model parameters, we exploit the redundancy in the space of $\boldsymbol{\mu}$ in terms of its impact on the policy. For example, increasing the mass of a limb will cause a similar effect on the optimal policy to increasing the torque limit of the motor connected to it. Therefore, we  learn a projection model that maps $\boldsymbol{\mu}$ down to a lower-dimensional latent variable $\boldsymbol{\eta} \in \mathbb{R}^M$, where $M$ is the dimension of the latent space ($M = 3$ in our experiments). We then condition the control policy directly on $\boldsymbol{\eta}$, instead of $\boldsymbol{\mu}$. We connect the last layer of projection module to the policy's input layer via a tanh activation such that the weights of both the projection module and the policy are trained together using the policy learning algorithm PPO \cite{schulman2017proximal}. This results in a policy that can exhibit different behaviors, modulated by $\boldsymbol{\eta}$. We call such policy a Projected Universal Policy (PUP): $\pi_{pup}: (\mathbf{s}, \boldsymbol{\eta}) \mapsto \mathbf{a}$.

\section{Post-training System Identification (Post-sysID)}
\label{sec:post-sysid}
During post-training system identification (post-sysID), we search in the space of $\boldsymbol{\eta}$ for the optimal $\boldsymbol{\eta}^*$ such that the conditioned policy $\pi_{pup}(\mathbf{s}; \boldsymbol{\eta}^*)$ achieves the best performance on the real hardware. Yu \etal \cite{yu2018policy} used CMA-ES for optimizing UP that achieves the best transfer performance and showed that it works comparably to Bayesian Optimization (BO). We use BO in this work because we found that for our problem with a low-dimensional $\boldsymbol{\eta}$, BO is in general more sample-efficient than CMA. 

The process of post-sysID starts with uniformly drawing $5$ samples in the space of $\boldsymbol{\eta}$ to build the initial Gaussian Process (GP) model for BO. For each sampled $\boldsymbol{\eta}$, we run the corresponding policy $\pi_{pup}$ on the robot to generate one trajectory and record the distance it travels before the robot loses balance. Note that the fitness function in post-sysD needs not to be the same as the reward function used for learning $\pi_{pup}$. We chose the simplest possible fitness function that only measures the distance travelled at the end of each rollout, but it can be easily replaced by more sophisticated fitness functions if necessary. At each iteration of BO, we use the latest GP model to find the next sample point that trades off between exploiting area near a previously good sample and exploring area that the GP model is uncertain about. We run BO for $20$ iterations and use the best $\boldsymbol{\eta}$ seen during optimization as the final output. The policy transfer process requires $25$ trajectories on the hardware, and the entire process takes less than $15$ minutes.



\section{Experiment}
\label{sec:results}

\begin{figure*}[t!]
\centering
\subfigure{\includegraphics[width=0.88\linewidth]{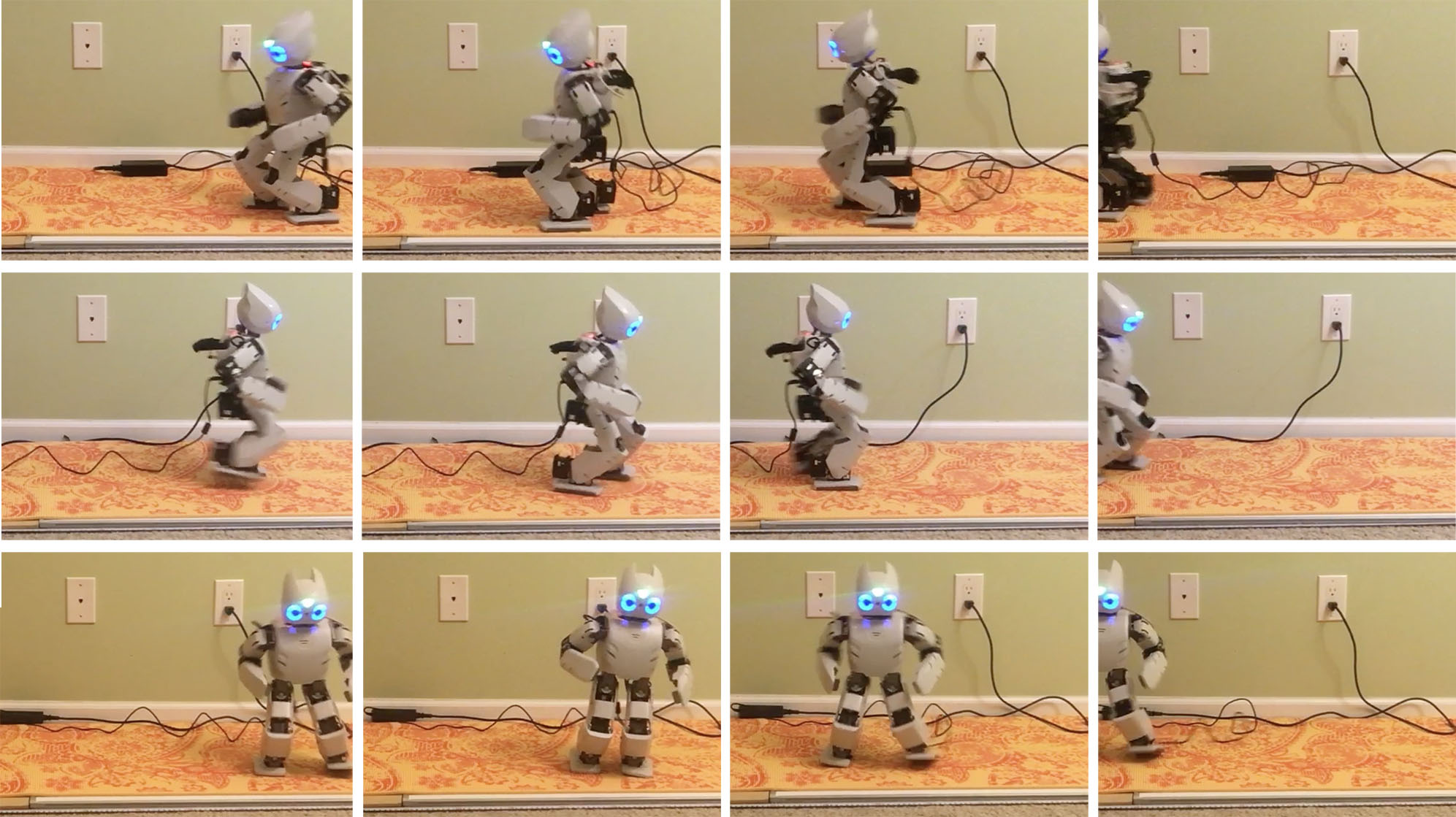}}
\caption{Illustration of locomotion policies deployed on the Darwin OP2 robot. Top: walk forward. Middle: walk backward. Bottom: walk sideways.}
\label{fig:result}
\vspace{-3mm}
\end{figure*}

\subsection{Experiment Setup}

\label{sec:exp_setup}

We test our algorithm on the Robotis Darwin OP2 robot. Darwin OP2 has $20$ Dynamixel MX-28T servo motors in total, $2$ on the head, $6$ on the arms and $12$ on the legs, all of which are controlled using target positions through a PID controller. We set the P gain, I gain and D gain of all the motors to be $32$, $0$ and $16$ on the hardware. Note that the PID controller for the actual motor is defined at the pulse width modulation (PWM) level, while the PD controller used in the simulation is defined at the torque level. Thus the gains used on the hardware is not transferable to the simulation. MX-28T provides decent sensing accuracy for the position and velocity of the motor. However, reading the position or velocity from all the motors takes about $10$ms, which limits our control frequency when both data are used. In this work, we instead use only the positions from the motors, and provide two consecutive motor position readings to the policy to provide information about the velocities. Darwin OP2 is also equipped with an on-board IMU sensor that provides raw measurements of angular velocity and linear acceleration of the robot torso. In order to have good estimation of the orientation or the robot, we need to collect and integrate data from the IMU sensor at high frequency. However, this is not possible because the IMU and the motors share the communication port. Therefore, instead of the on-board IMU, we use the Bosch bno055 IMU sensor for estimating the orientation of the robot. With this augmentation, we can reach a control frequency of $33Hz$. We use a physics-based simulator, Dart \cite{lee2018dart}, to simulate the robot's behavior under different control signals.

To evaluate our approach, we train locomotion policies that control the Darwin OP2 robot to walk forward, backward and sideways in simulation and transfer to the real hardware. We first collect motion trajectories of the real robot performing manual-scripted movements. For each motor on the robot, we apply a step function action starting from a random pose and record their responses while suspending the robot to avoid ground contact. One such example can be seen in Figure \ref{fig:sysid_comp}. We use step functions of magnitude $0.1$, $0.3$ and $0.6$ to collect motor behaviors at different speeds. In addition, we also design trajectories where the robot stands up and falls in different directions. For trajectories that involve ground contact, we also record the estimated orientation from the IMU sensor. The set of movements we use can be seen in the supplementary video \footnote{https://youtu.be/bq8xZgbLHcw}.

The robot is tasked to walk on a yoga mat that lies on top of a white board, as shown in Figure \ref{fig:result}. We choose this deformable surface to better provide protection for the robot. The performance of the policy can be viewed in the supplementary video.

For all examples in this paper, the state space includes the position of motors and the estimated orientation represented in Euler angles for two consecutive timesteps. The action space is defined as the target positions for each motor. To accelerate the learning process in simulation, we use a reference trajectory of robot stepping in place that was generated manually. The action of the policy is to apply adjustment to the reference trajectory:  
\begin{equation}
\vc{q}_{target} = \vc{q}_{ref} + \delta  \pi_{pup}(\vc{s}; \boldsymbol{\eta^*}), \nonumber
\end{equation}
 where $\delta$ controls the magnitude of the adjustment and the policy $\pi_{pup}$ outputs a value in $[-1, 1]$. We use $\delta=0.3$ for walking forward and sideways and $\delta=0.2$ for walking backwards. Note that the reference trajectory does not need to be dynamically feasible, as tracking our stepping-in-place reference trajectory causes the robot to fall immediately. Similar to \cite{learndexmanipulation18}, we also discretize the action space into $11$ bins in each dimension to further accelerate the policy training.

We use Proximal Policy Optimization (PPO) to train the control policies and use the following reward function:
\begin{equation}
r(\vc{s},\vc{a})= w_v E_v + w_a E_a + w_w E_w + w_t E_t + E_c. \nonumber
\end{equation}
The first term $E_v = \| \dot{\mathbf{x}}\|$ encourages the robot to move as fast as possible in the direction $\mathbf{x}$. The second and third term $E_a = ||\boldsymbol{\tau}||^2$, $E_w = \boldsymbol{\tau}\cdot \dot{\vc{q}}$ penalize the torque and work applied to the motors, where $\boldsymbol{\tau}$ is the resulting torque applied to the motor under the action $\vc{a}$.  $E_t=||\vc{q}^t-\vc{q}^t_{ref}||^2$ rewards the robot to track the reference trajectory, where $\vc{q}^t_{ref}$ denotes the reference trajectory at timestep $t$. Finally, $E_c=5$ is a constant reward for not falling to the ground. We use an identical reward function with $w_v=10.0, w_a=0.01$, $w_w=0.005$, $w_t=0.2$ for all of the presented examples. We also use the mirror symmetry loss proposed in \cite{YuSIGGRAPH2018} during training of PUP, which we found to improve the quality of the learned locomotion gaits. For controlling the robot to walk in different directions, we rotate the robot's coordinate frame such that the desired walking direction is aligned with the positive $x$-axis in the robot frame.

\subsection{Baselines}
We evaluate our method by comparing it with two baselines. For the first baseline, we optimize for a single model $\boldsymbol{\mu}$ during Pre-sysID instead of a range of $\boldsymbol{\mu}$, and use the model to train a policy. We denote this baseline ``Nominal''. The second baseline uses the range of $\boldsymbol{\mu}$ computed by Pre-sysID and trains a robust policy through domain randomization with that range. We denote the second baseline ``Robust''.

To account for uncertainty in the sensors and networking, we model additional noise in the simulation during policy training for our method and the two baselines. Specifically, we randomly set the control frequency to be in $[25, 33]$Hz for each rollout, add a bias to the estimated orientation drawn from $\mathcal{U}(-0.3, 0.3)$ and add a Gaussian noise of standard deviation $0.01$ to the observed motor position. In addition, we add noise drawn from $\mathcal{U}(-0.25, 0.25)$ to the input $\boldsymbol{\mu}$ during the training of PUP to improve the robustness of the policy.

\subsection{Performance on Locomotion Tasks}

\begin{figure}[t!]
\centering
\subfigure{\includegraphics[width=0.9\linewidth]{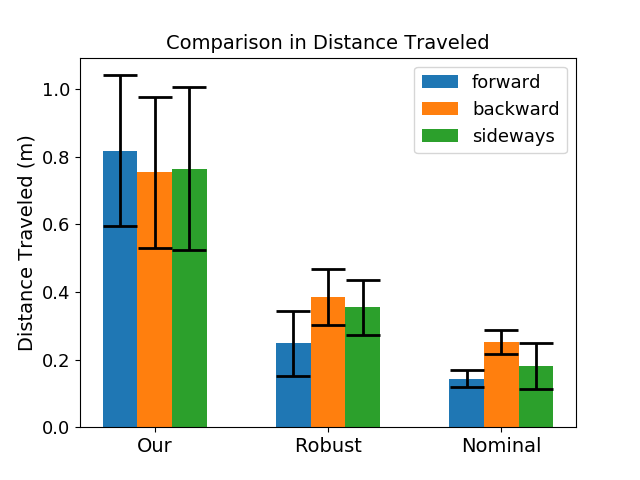}}
\vspace{-2mm}
\caption{Comparison of the distance travelled by the robot using our method and the baselines. Error bars indicate one standard deviation from five runs of the same policy.}
\label{fig:fbs_stat}
\vspace{-3mm}
\end{figure}
\begin{figure}[t!]
\centering
\subfigure{\includegraphics[width=0.9\linewidth]{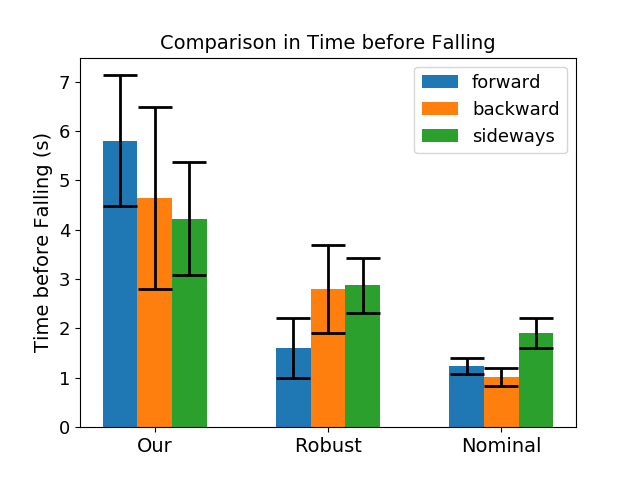}}
\vspace{-2mm}
\caption{Comparison of the elapsed time before the robot loses balance using our method and baselines. Error bars indicate one standard deviation from five runs of the same policy.}
\label{fig:fbs_stat_fall}
\vspace{-3mm}
\end{figure}

Figure \ref{fig:fbs_stat} and Figure \ref{fig:fbs_stat_fall} show the comparison between our method and the baselines. We evaluate a trained policy by running it $5$ times on the real hardware, and measure the distance and time before the robot loses balance or reaches the end of the power cable. Our method clearly outperforms the baselines and is the only method that can control the robot to walk to, and occasionally beyond, the edge of the white board (at $0.8$m). Because PUP is trained to be specialized for different environments, it learns to take larger steps than Robust, which tends to take conservative actions. This results in a faster walking gait, and may also have contributed to the larger variance seen in our policy. An illustration of the three locomotion tasks with our trained policies can be seen in Figure \ref{fig:result}.

\subsection{Effect of using NN-PD Actuators}

\begin{figure}[t!]
\centering
\subfigure{\includegraphics[width=5.5cm]{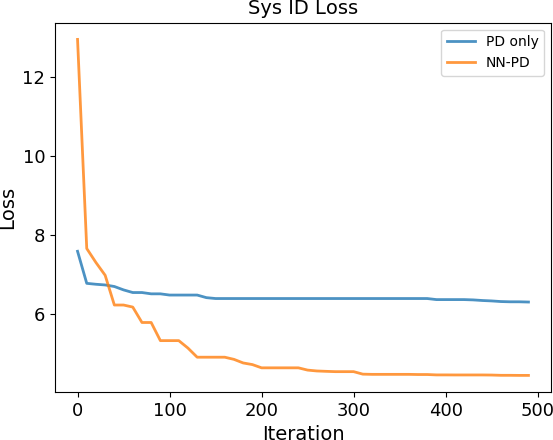}}
\caption{Comparison of system identification performance with and without NN-PD actuators. Both models are optimized using the same set of real-world data and the reported loss is calculated according to Equation \ref{eq:sysid_loss}.}
\label{fig:sysid_loss_comp}
\end{figure}

\begin{figure}[t!]
\centering
\subfigure{\includegraphics[width=\linewidth]{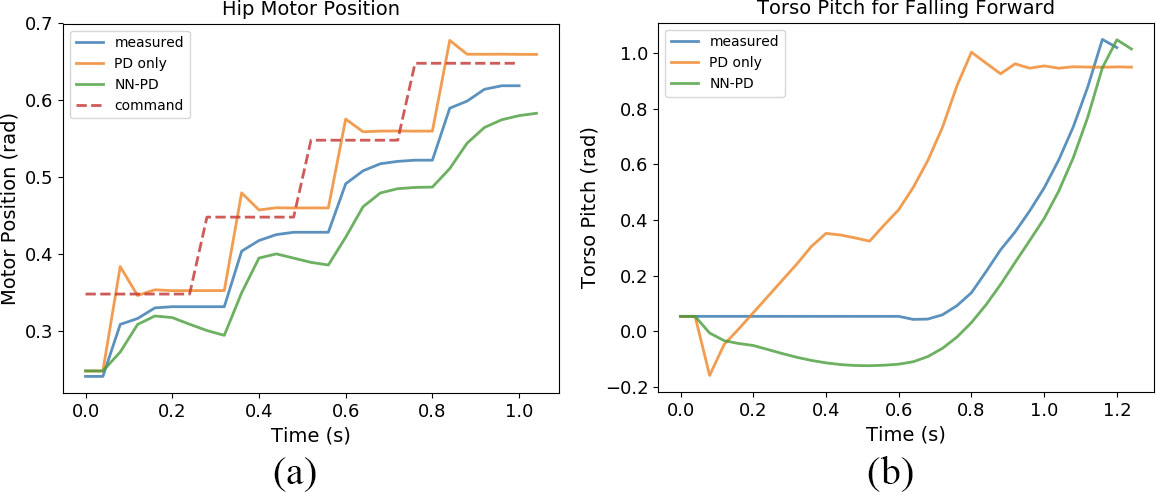}}
\vspace{-3mm}
\caption{System identification performance comparison of NN-PD and PD only on (a) the hip motor position during step function command with magnitude $0.1$ and (b) torso pitch during falling forward motion.}
\label{fig:sysid_comp}
\end{figure}

Our method models the motor dynamics as a neural network paired with a PD controller. Here we examine the necessity of having this additional components in the model. Specifically, we identify two models, one with NN-PD controllers and one without (PD only), using the same set of real-world data. Figure \ref{fig:sysid_loss_comp} shows the the optimization curve over $500$ iterations for both models. We can see that NN-PD is able to achieve a notably better loss compared to PD only. To further demonstrate the behavior of the two models, we plot the simulated motor position for the hip joint when a step function is applied, and the estimated pitch of the torso when a trajectory controls the robot to fall forward, as shown in Figure \ref{fig:sysid_comp} (a) and (b). We can see that NN-PD is able to better reproduce the overall behavior than PD only.

\subsection{Identified Model Parameter Bounds}

\begin{figure}[t!]
\centering
\subfigure{\includegraphics[width=7.5cm]{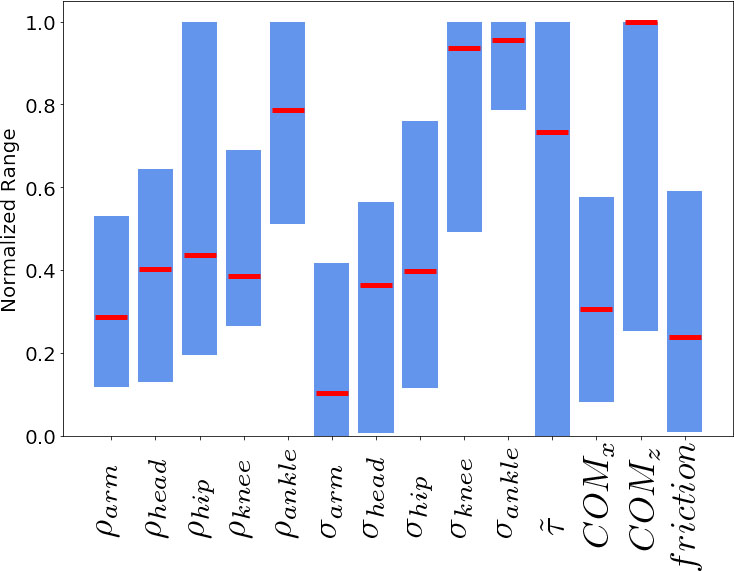}}
\caption{Identified model parameter bounds (blue bars) and the nominal parameters (red lines).}
\label{fig:simparam_range}
\end{figure}

Figure \ref{fig:simparam_range} visualizes  $\boldsymbol{\mu}_{lb}$ and $\boldsymbol{\mu}_{ub}$ identified by the pre-sysID stage. We normalize the search range of each model parameter to be in $[0, 1]$ and show the identified bounds as the blue bars. The red lines indicate the nominal parameters $\hat{\boldsymbol{\mu}}$ optimized using the entire set of pre-sysID trajectories. Some parameters, such as $\sigma_{ankle}$, have tighter bounds, which indicate higher confidence in the optimized values for those parameters. The parameters with wider range indicate that no single value of $\boldsymbol{\mu}$ can explain all the training trajectories well and naively using the nominal values for these parameters may lead to poor transfer performance. One example of such parameters is the bounds for $\tilde{\tau}$. Upon further examination, we found that the identified $\tilde{\tau}$ tends to be bipolar depending on whether the subset of training trajectories involves contact or not. These phenomena suggest that our current model is still not expressive enough to explain all the real-world observations and further improvement in modeling may be necessary for transferring more challenging tasks. We also note that, partially due to the wide range of motions for pre-sysID, the identified bounds are not necessarily useful for the tasks of interest. For example, the parameters associated with the head, $\rho_{head}$ and $\sigma_{head}$, have wide bounds but their impact to locomotion tasks is relatively small. During the training of PUP, the projection module will learn to ignore the variations in these parameters. 

\section{Discussion and Conclusion}

We have proposed a transfer learning algorithm for learning robotic controllers in physics simulation and applying them to the real hardware. The key idea in our method is to perform system identification in two stages connected by a Projected Universal Policy, whose behavior is modulated by a low dimensional latent variable. We demonstrate our method on training locomotion policies for the Darwin OP2 robot to walk forward, backward and sideways and transfer to the real robot using $25$ trials on the hardware.

During post-sysID, our method uses additional task-relevant data to help identify the optimal conditioned policy $\pi_{pup}(\mathbf{s}; \boldsymbol{\eta}^*)$. To provide a fair comparison, one could also use task-relevant data to further improve the baseline policies. However, to fully take advantage of these trajectories for policy learning, one would need to upgrade sensor instrumentation for measuring global position and orientation needed in reward function evaluation. In contrast, our method only uses these trajectories for post-sysID with very simple fitness function that only measures the traveling distance and the elapsed time. In addition, the size of the task-relevant data (less than $2500$ steps) is only enough to perform one iteration of PPO in a typical setting. In comparison, we use $20,000$ steps per learning iteration in simulation. For those reasons, we do not believe that such a small amount of task-relevant data can further improve the results of baseline methods in our experiments.


The set of model parameters $\boldsymbol{\mu}$ used in our work is currently chosen manually based on prior knowledge about the robot and the tasks of interest. Although it works for Darwin OP2 learning locomotion tasks, it is not clear as to how well it can be applied to different robotic hardware or different tasks, such as picking up objects or climbing ladders. Investigating a systematic and automatic way to select model parameters would be an interesting future research direction.

\addtolength{\textheight}{-0cm}   




\section*{ACKNOWLEDGMENT}
We thank Charles C. Kemp, Alex Clegg,
Zackory Erickson and Henry M. Clever for the insightful discussions. This work was supported by NSF award IIS-1514258 and AWS Cloud Credits for Research.

\bibliographystyle{IEEEtran}
\bibliography{references}

\end{document}